\documentclass[10pt]{article}
\newcommand\tab[1][1cm]{\hspace*{#1}}
\usepackage{footmisc}
\usepackage[utf8]{inputenc}
\usepackage{microtype}
\usepackage{graphicx}
\usepackage{subfigure}
\usepackage{booktabs} 
\usepackage{hyperref}

\usepackage[accepted]{icml2020}

\icmltitlerunning{Detecting Potential Topics In News Using BERT, CRF and Wikipedia}

\begin{document}

\twocolumn[
\icmltitle{Detecting Potential Topics In News Using BERT, CRF and Wikipedia}

\icmlsetsymbol{equal}{}

\begin{icmlauthorlist}
\icmlauthor{Swapnil Ashok Jadhav,}{}
\icmlauthor{Dailyhunt}{}
\end{icmlauthorlist}

\icmlsetsymbol{equal}{}

\vskip 0.3in
]

\begin{abstract}
For a news content distribution platform like Dailyhunt\footnote{\url{https://www.dailyhunt.com}}, Named Entity Recognition is a pivotal task for building better user recommendation and notification algorithms. Apart from identifying names, locations, organisations from the news for 13+ Indian languages and use them in  algorithms, we also need to identify n-grams which do not necessarily fit in the definition of Named-Entity, yet they are important. For example, {\it "me too movement", "beef ban", "alwar mob lynching"}. In this exercise, given an English language text, we are trying to detect case-less n-grams  which convey important information and can be used as topics and/or hashtags for a news. Model is built using Wikipedia titles data, private English news corpus and BERT-Multilingual\cite{devlin2018bert} pre-trained model, Bi-GRU\cite{2014arXiv1412.3555C} and CRF architecture. It shows promising results when compared with industry best Flair\footnote{\url{https://github.com/flairNLP/flair}}, Spacy\footnote{\url{https://spacy.io/api/entityrecognizer}} and Stanford-caseless-NER\footnote{\url{https://stanfordnlp.github.io/CoreNLP/caseless.html}} in terms of F1 and especially Recall.
\end{abstract}

\section{Introduction \& Related Work}

Named-Entity-Recognition(NER) approaches can be categorised broadly in three types. Detecting NER with predefined dictionaries and rules\cite{Florian11}, with some statistical approaches\cite{Ratinov11} and with deep learning approaches\cite{Dong11}. 


Stanford CoreNLP NER is a widely used baseline for many applications \cite{Stanford11}. Authors have used approaches of Gibbs sampling and conditional random field (CRF) for non-local information gathering and then Viterbi algorithm to infer the most likely state in the CRF sequence output\cite{Finkel11}.

Deep learning approaches in NLP use document, word or token representations instead of one-hot encoded vectors. With the rise of transfer learning, pretrained Word2Vec\cite{2013arXiv1301.3781M}, GloVe\cite{glove11}, fasttext\cite{fasttext11} which provides word embeddings were being used with recurrent neural networks (RNN) to detect NERs. Using LSTM layers followed by CRF layes with pretrained word-embeddings as input has been explored here\cite{2015arXiv150801991H}. Also, CNNs with character embeddings as inputs followed by bi-directional LSTM and CRF layers, were explored here\cite{2016arXiv160301354M}. 

With the introduction of attentions and transformers\cite{2017arXiv170603762V} many deep architectures emerged in last few years. Approach of using these pretrained models like Elmo\cite{2018arXiv180205365P}, Flair\cite{akbik-etal-2019-flair} and BERT\cite{devlin2018bert} for word representations followed by variety of LSMT and CRF combinations were tested by authors in \cite{2019arXiv190806926S} and these approaches show  state-of-the-art performance. 

There are very few approaches where caseless NER task is explored. In this recent paper\cite{2019arXiv190311222M} authors have explored effects of "Cased" entities and how variety of networks perform and they show that the most effective strategy is a concatenation of cased and lowercased training data, producing a single model with high performance on both cased and uncased text. 

In another paper\cite{2019arXiv191207095M}, authors have proposed True-Case pre-training before using BiLSTM+CRF approach to detect NERs effectively. Though it shows good results over previous approaches, it is not useful in Indian Languages context as there is no concept of cases.

In our approach, we are focusing more on data preparation for our definition of topics using some of the state-of-art architectures based on BERT, LSTM/GRU and CRF layers as they have been explored in previous approaches mentioned above. Detecting caseless topics with higher recall and reasonable precision has been given a priority over f1 score. And comparisons have been made with available and ready-to-use open-source libraries from the productionization perspective.


\section{Data Preparation}
We need good amount of data to try deep learning state-of-the-art algorithms.
There are lot of open datasets\footnote{\url{https://github.com/juand-r/entity-recognition-datasets}} available for names, locations, organisations, but not for topics as defined in Abstract above. Also defining and inferring topics is an individual preference and there are no fix set of rules for its definition. But according to our definition, we can use wikipedia titles as our target topics. English wikipedia dataset\footnote{\url{https://dumps.wikimedia.org/enwiki/}} has more than 18 million titles if we consider all versions of them till now. We had to clean up the titles to remove junk titles as wikipedia title almost contains all the words we use daily. To remove such titles, we deployed simple rules as follows - 

\begin{itemize}
    \item Remove titles with common words : "are", "the", "which"
    \item Remove titles with numeric values : 29, 101
    \item Remove titles with technical components, driver names, transistor names : X00, lga-775
    \item Remove 1-gram titles except locations (almost 80\% of these also appear in remaining n-gram titles)
\end{itemize}

After doing some more cleaning we were left with ~10 million titles. We have a dump of ~15 million English news articles published in past 4 years. Further, we reduced number of articles by removing duplicate and near similar articles. We used our pre-trained doc2vec models and cosine similarity to detect almost similar news articles. Then selected minimum articles required to cover all possible 2-grams to 5-grams. This step is done to save some training time without loosing accuracy. Do note that, in future we are planning to use whole dataset and hope to see gains in F1 and Recall further. But as per manual inspection, our dataset contains enough variations of sentences with rich vocabulary which contains names of celebrities, politicians, local authorities, national/local organisations and almost all locations, India and International, mentioned in the news text, in last 4 years.

We then created a parallel corpus format as shown  in Table 1. Using pre-trained Bert-Tokenizer\footnote{\url{https://pypi.org/project/pytorch-pretrained-bert/}} from hugging-face, converted words in sentences to tokenes. Caseless-BERT pre-trained tokenizer is used. Notice that some of the topic words are broken into tokens and {\it NER} tag has been repeated accordingly. For example, in Table 1 second row, word \textit{"harassment"} is broken into \textit{"har \#\#ass \#\#ment"}. Similarly, one "NER" tag is repeated three times to keep the length of sequence-pair same. Finally, for around 3 million news articles, parallel corpus is created, which is of around 150 million sentences, with around 3 billion words (all lower cased) and with around 5 billion tokens approximately.

\begin{table}
\centering
  \caption{Parallel Corpus Preparation with BERT Tokenizer}
  \label{tab:freq}
  \begin{tabular}{p{1.75cm}p{6.0cm}}
    \toprule
    \textbf{Text} & the me too movement with a large variety of local and international related names , is a movement against sexual harassment and sexual assault \\
    \textbf{NER Tags} & 0 NER NER NER 0 0 0 0 0 0 0 0 0 0 0 0 0 0 0 0 NER NER 0 NER NER \\
    \midrule
    \textbf{Tokenized Text} & the me too movement with a large variety of local and international related names, is a movement against sexual har \#\#ass \#\#ment and sexual assault \\
    \textbf{Tokenized NER Tags} & 0 NER NER NER 0 0 0 0 0 0 0 0 0 0 0 0 0 0 0 0 NER NER NER NER 0 NER NER \\
  \bottomrule
\end{tabular}
\end{table}

\section{Experiments}
\subsection{Model Architecture}
We tried multiple variations of LSTM and GRU layes, with/without CRF layer. There is a marginal gain in using GRU layers over LSTM. Also, we saw gain in using just one layers of GRU instead of more. Finally, we settled on the architecture, shown in Figure 1 for the final training, based on validation set scores with sample training set. 

Text had to be tokenized using pytorch-pretrained-bert as explained above before passing to the network. Architecture is built using tensorflow/keras. Coding inspiration taken from BERT-keras\footnote{\url{https://github.com/Separius/BERT-keras}} and for CRF layer  keras-contrib\footnote{\url{https://github.com/keras-team/keras-contrib/}}. If one is more comfortable in pytorch there are many examples available on github, but pytorch-bert-crf-ner\footnote{\url{https://github.com/eagle705/pytorch-bert-crf-ner}} is better for an easy start.

We used BERT-Multilingual model so that we can train and fine-tune the same model for other Indian languages. You can take BERT-base or BERT-large for better performance with only English dataset. Or you can use DistilBERT for English and DistilmBERT for 104 languages  \footnote{\url{https://github.com/huggingface/transformers/tree/master/examples/distillation}} for faster pre-training and inferences. Also, we did not choose AutoML approach for hyper-parameter tuning which could have resulted in much more accurate results but at the same time could have taken very long time as well. So instead, chose and tweaked the parameters based on initial results.

We trained two models, one with sequence length 512 to capture document level important n-grams and second with sequence length 64 to capture sentence/paragraph level important n-grams. Through experiments it was evident that, sequence length plays a vital role in deciding context and locally/globally important n-grams. Final output is a concatenation of both the model outputs.

\begin{figure}[ht]
  \centering
  \includegraphics[width=200pt, height=250pt]{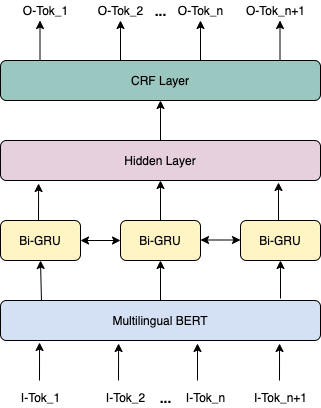} %
  \caption{BERT + Bi-GRU + CRF, Final Architecture Chosen For Topic Detection Task.}
\end{figure}

\subsection{Training}
Trained the topic model on single 32gb NVidia-V100 and it took around 50 hours to train the model with sequence length 512. We had to take 256gb ram machine to accommodate all data in memory for faster read/write. Also, trained model with 64 sequence length in around 17 hours.

It is very important to note that sequence length decides how many bert-tokens you can pass for inference and also decides training time and accuracy. Ideally more is better because inference would be faster as well. For 64 sequence length, we are moving 64-token window over whole token-text and recognising topics in each window. So, one should choose sequence length according to their use case. Also, we have explained before our motivation of choosing 2 separate sequence lengths models.

We stopped the training for both the models when it crossed 70\% precision, 90\% recall on training and testing sets, as we were just looking to get maximum recall and not bothered about precision in our case. Both the models reach this point at around 16 epochs.

\subsection{Results}
Comparison with existing open-source NER libraries is not exactly fair as they are NOT trained for detecting topics and important n-grams, also NOT trained for case-less text. But they are useful in testing and benchmarking if our model is detecting traditional NERs or not, which it should capture, as Wikipedia titles contains almost all Names, Places and Organisation names. You can check the sample output here\footnote{\url{https://github.com/swapniljadhav1921/bert\_crf\_topic\_detection/}} 

Comparisons have been made among Flair-NER, Stanford-caseless-NER (used english.conll.4class.caseless as it performed better than 3class and 7class), Spacy-NER and our models. Of which only Stanford-NER provides case-less models. In Table 2, scores are calculated by taking traditional NER list as reference. In Table 4, same is done with Wikipedia Titles reference set. 

As you can see in Table 2 \& 3, recall is great for our model but precision is not good as Model is also trying to detect new potential topics which are not there even in reference Wikipedia-Titles and NER sets. In capturing Wikipedia topics our model clearly surpasses other models in all scores. 

Spacy results are good despite not being trained for case-less data. In terms of F1 and overall stability Spacy did better than Stanford NER, on our News Validation set. Similarly, Stanford did well in Precision but could not catch up with Spacy and our model in terms of Recall.
Flair overall performed poorly, but as said before these open-source models are not trained for our particular use-case.

\begin{table}
\centering
\caption{Comparison with Traditional NERs as reference}
\begin{tabular}{lccr}  
\toprule
Models & Precision & Recall & F1 \\
\midrule
BERT+BiGRU+CRF & 60.09 & \textbf{80.08} & \textbf{68.66} \\
Stanford & \textbf{90.54} & 37.17 & 52.70 \\
Spacy & 88.71 & 55.05 & \textbf{67.94} \\
Flair & 85.62 & 10.28 & 18.36 \\
\bottomrule
\end{tabular}
\label{tab:booktab1}
\end{table}

\begin{table}
\centering
\caption{Comparison with Wikipedia titles as reference}
\begin{tabular}{lccr}  
\toprule
Models & Precision & Recall & F1 \\
\midrule
BERT+BiGRU+CRF & 51.97 & \textbf{69.76} & \textbf{59.56} \\
Stanford & 52.83 & 19.88 & 28.89 \\
Spacy & 36.31 & 26.40 & 30.57 \\
Flair & 65.36 & 7.33 & 13.17 \\
\bottomrule
\end{tabular}

\label{tab:booktab11} 
\end{table}

\subsection{Discussions}
Lets check some examples for detailed analysis of the models and their results. Following is the economy related news.

\textbf{Example 1 : }\textit{\footnotesize around \$1–1.5 trillion or around two percent of global gdp, are lost to corruption every year, president of the natural resource governance institute nrgi has said. speaking at a panel on integrity in public governance during the world bank group and international monetary fund annual meeting on sunday, daniel kaufmann, president of nrgi, presented the statistic, result of a study by the nrgi, an independent, non-profit organisation based in new york. however, according to kaufmann, the figure is only the direct costs of corruption as it does not factor in the opportunities lost on innovation and productivity, xinhua news agency reported. a country that addresses corruption and significantly improves rule of law can expect a huge increase in per capita income in the long run, the study showed. it will also see similar gains in reducing infant mortality and improving education, said kaufmann.}

Detected NERs can be seen per model in Table 4. Our model do not capture numbers as we have removed all numbers from my wiki-titles as topics. Reason behind the same is that we can easily write regex to detect currency, prices, time, date and deep learning is not required for the same. Following are few important n-grams only our models was able to capture -  

\tab\textit{\footnotesize capita income}\\
\tab\textit{\footnotesize infant mortality}\\
\tab\textit{\footnotesize international monetary fund annual meeting}\\
\tab\textit{\footnotesize natural resource governance institute}\\
\tab\textit{\footnotesize public governance}

At the same time, we can see that Spacy did much better than Stanford-caseless NER and Flair could not capture any of the NERs. Another example of a news in political domain and detected NERs can be seen per model in Table 5.

\textbf{Example 2 : } \textit{\footnotesize wearing the aam aadmi party's trademark cap and with copies of the party's five-year report card in hand, sunita kejriwal appears completely at ease. it's a cold winter afternoon in delhi, as the former indian revenue service (irs) officer hits the campaign trail to support her husband and batchmate, chief minister arvind kejriwal. emerging from the background for the first time, she is lending her shoulder to the aap bandwagon in the new delhi assembly constituency from where the cm, then a political novice, had emerged as the giant killer by defeating congress incumbent sheila dikshit in 2013.}

Correct n-grams captured only by our model are -

\tab \textit{\footnotesize aam aadmi party} \\
\tab \textit{\footnotesize aap bandwagon} \\
\tab \textit{\footnotesize delhi assembly constituency} \\
\tab \textit{\footnotesize giant killer} \\
\tab \textit{\footnotesize indian revenue service} \\
\tab \textit{\footnotesize political novice}

In this example, Stanford model did better and captured names properly, for example {\it "sheila dikshit"} which Spacy could not detect but Spacy captureed almost all numeric values along with numbers expressed in words.

It is important to note that, our model captures NERs with some additional words around them. For example, "president of nrgi" is detected by the model but not "ngri". But model output does convey more information than the later. To capture the same for all models (and to make comparison fair), partial match has been enabled and if correct NER is part of predictied NER then later one is marked as matched. This could be the reason for good score for Spacy. Note that, partial match is disabled for Wikipedia Titles match task as shown in Table 3. Here, our model outperformed all the models.

\section{Conclusion and Future Work}
Through this exercise, we were able to test out the best suitable model architecture and data preparation steps so that similar models could be trained for Indian languages. Building cased or caseless NERs for English was not the final goal and this has already been benchmarked and explored before in previous approaches explained in "Related Work" section. We  didn't use traditional datasets for model performance comparisons \& benchmarks. As mentioned before, all the comparisons are being done with open-source models and libraries from the productionization point of view. We used a english-news validation dataset which is important and relevant to our specific task and all validation datasets and raw output results can be found at our github link \footnote{\url{https://github.com/swapniljadhav1921/bert\_crf\_topic\_detection}}.

Wikipedia titles for Indian languages are very very less and resulting tagged data is even less to run deep architectures. We are trying out translations/transliterations of the English-Wiki-Titles to improve Indic-languages entity/topics data.

This approach is also useful in building news-summarizing models as it detects almost all important n-grams present in the news. Output of this model can be introduced in a summarization network to add more bias towards important words and bias for their inclusion.

\bibliography{example_paper}

\begin{thebibliography}{18}
\providecommand{\natexlab}[1]{#1}
\providecommand{\url}[1]{\texttt{#1}}
\expandafter\ifx\csname urlstyle\endcsname\relax
  \providecommand{\doi}[1]{doi: #1}\else
  \providecommand{\doi}{doi: \begingroup \urlstyle{rm}\Url}\fi

\bibitem[Akbik et~al.(2019)Akbik, Bergmann, Blythe, Rasul, Schweter, and
  Vollgraf]{akbik-etal-2019-flair}
Akbik, A., Bergmann, T., Blythe, D., Rasul, K., Schweter, S., and Vollgraf, R.
\newblock {FLAIR}: An easy-to-use framework for state-of-the-art {NLP}.
\newblock In \emph{Proceedings of the 2019 Conference of the North {A}merican
  Chapter of the Association for Computational Linguistics (Demonstrations)},
  pp.\  54--59, Minneapolis, Minnesota, June 2019. Association for
  Computational Linguistics.
\newblock \doi{10.18653/v1/N19-4010}.
\newblock URL \url{https://www.aclweb.org/anthology/N19-4010}.

\bibitem[Christopher~Manning \& McClosky(2014)Christopher~Manning and
  McClosky]{Stanford11}
Christopher~Manning, Mihai~Surdeanu, J. B. J. F. S.~B. and McClosky, D.
\newblock The stanford corenlp natural language processing toolkit.
\newblock \emph{Proceedings of 52nd Annual Meeting of the Association for
  Computational Linguistics: System Demonstrations, 2014.}, 2014.

\bibitem[{Chung} et~al.(2014){Chung}, {Gulcehre}, {Cho}, and
  {Bengio}]{2014arXiv1412.3555C}
{Chung}, J., {Gulcehre}, C., {Cho}, K., and {Bengio}, Y.
\newblock {Empirical Evaluation of Gated Recurrent Neural Networks on Sequence
  Modeling}.
\newblock \emph{arXiv e-prints}, art. arXiv:1412.3555, Dec 2014.

\bibitem[Devlin et~al.(2018)Devlin, Chang, Lee, and Toutanova]{devlin2018bert}
Devlin, J., Chang, M.-W., Lee, K., and Toutanova, K.
\newblock Bert: Pre-training of deep bidirectional transformers for language
  understanding.
\newblock \emph{arXiv preprint arXiv:1810.04805}, 2018.

\bibitem[{Huang} et~al.(2015){Huang}, {Xu}, and {Yu}]{2015arXiv150801991H}
{Huang}, Z., {Xu}, W., and {Yu}, K.
\newblock {Bidirectional LSTM-CRF Models for Sequence Tagging}.
\newblock \emph{arXiv e-prints}, art. arXiv:1508.01991, Aug 2015.

\bibitem[Jeffrey~Pennington \& Manning.(2014)Jeffrey~Pennington and
  Manning.]{glove11}
Jeffrey~Pennington, R.~S. and Manning., C.
\newblock Glove: Global vectors for word representation.
\newblock \emph{Proceedings of the 2014 Conference on Empirical Methods in
  Natural Language Processing (EMNLP)}, 2014.

\bibitem[Jenny Rose~Finkel \& Manning(2005)Jenny Rose~Finkel and
  Manning]{Finkel11}
Jenny Rose~Finkel, T.~G. and Manning, C.
\newblock Incorporating non-local information into information extraction
  systems by gibbs sampling.
\newblock \emph{Proceedings of the 43rd Annual Meeting on Association for
  Computational Linguistics - ACL ’05}, 2005.

\bibitem[{Ma} \& {Hovy}(2016){Ma} and {Hovy}]{2016arXiv160301354M}
{Ma}, X. and {Hovy}, E.
\newblock {End-to-end Sequence Labeling via Bi-directional LSTM-CNNs-CRF}.
\newblock \emph{arXiv e-prints}, art. arXiv:1603.01354, Mar 2016.

\bibitem[{Mayhew} et~al.(2019{\natexlab{a}}){Mayhew}, {Gupta}, and
  {Roth}]{2019arXiv191207095M}
{Mayhew}, S., {Gupta}, N., and {Roth}, D.
\newblock {Robust Named Entity Recognition with Truecasing Pretraining}.
\newblock \emph{arXiv e-prints}, art. arXiv:1912.07095, Dec 2019{\natexlab{a}}.

\bibitem[{Mayhew} et~al.(2019{\natexlab{b}}){Mayhew}, {Tsygankova}, and
  {Roth}]{2019arXiv190311222M}
{Mayhew}, S., {Tsygankova}, T., and {Roth}, D.
\newblock {ner and pos when nothing is capitalized}.
\newblock \emph{arXiv e-prints}, art. arXiv:1903.11222, Mar 2019{\natexlab{b}}.

\bibitem[{Mikolov} et~al.(2013){Mikolov}, {Chen}, {Corrado}, and
  {Dean}]{2013arXiv1301.3781M}
{Mikolov}, T., {Chen}, K., {Corrado}, G., and {Dean}, J.
\newblock {Efficient Estimation of Word Representations in Vector Space}.
\newblock \emph{arXiv e-prints}, art. arXiv:1301.3781, Jan 2013.

\bibitem[{Peters} et~al.(2018){Peters}, {Neumann}, {Iyyer}, {Gardner}, {Clark},
  {Lee}, and {Zettlemoyer}]{2018arXiv180205365P}
{Peters}, M.~E., {Neumann}, M., {Iyyer}, M., {Gardner}, M., {Clark}, C., {Lee},
  K., and {Zettlemoyer}, L.
\newblock {Deep contextualized word representations}.
\newblock \emph{arXiv e-prints}, art. arXiv:1802.05365, Feb 2018.

\bibitem[Piotr~Bojanowski \& Mikolov.(2017)Piotr~Bojanowski and
  Mikolov.]{fasttext11}
Piotr~Bojanowski, Edouard~Grave, A.~J. and Mikolov., T.
\newblock Enriching word vectors with subword information.
\newblock \emph{Transactions of the Association for Computational Linguistics,
  5:135–146}, 2017.

\bibitem[R.~Florian \& Zhang(2003)R.~Florian and Zhang]{Florian11}
R.~Florian, A.~Ittycheriah, H.~J. and Zhang, T.
\newblock Named entity recognition through classifier combination.
\newblock \emph{Proceedings of the seventh conference on Natural language
  learning at HLT-NAACL 2003-Volume 4, 2003: Association for Computational
  Linguistics}, 2003.

\bibitem[Ratinov \& Roth(2009)Ratinov and Roth]{Ratinov11}
Ratinov, L. and Roth, D.
\newblock Design challenges and misconceptions in named entity recognition.
\newblock \emph{Proceedings of the thirteenth conference on computational
  natural language learning, 2009: Association for Computational Linguistics},
  2009.

\bibitem[{Strakov{\'a}} et~al.(2019){Strakov{\'a}}, {Straka}, and
  {Haji{\v{c}}}]{2019arXiv190806926S}
{Strakov{\'a}}, J., {Straka}, M., and {Haji{\v{c}}}, J.
\newblock {Neural Architectures for Nested NER through Linearization}.
\newblock \emph{arXiv e-prints}, art. arXiv:1908.06926, Aug 2019.

\bibitem[{Vaswani} et~al.(2017){Vaswani}, {Shazeer}, {Parmar}, {Uszkoreit},
  {Jones}, {Gomez}, {Kaiser}, and {Polosukhin}]{2017arXiv170603762V}
{Vaswani}, A., {Shazeer}, N., {Parmar}, N., {Uszkoreit}, J., {Jones}, L.,
  {Gomez}, A.~N., {Kaiser}, L., and {Polosukhin}, I.
\newblock {Attention Is All You Need}.
\newblock \emph{arXiv e-prints}, art. arXiv:1706.03762, Jun 2017.

\bibitem[X.~Dong \& Yang(2016)X.~Dong and Yang]{Dong11}
X.~Dong, L.~Qian, Y. G. L. H. Q.~Y. and Yang, J.
\newblock A multiclass classification method based on deep learning for named
  entity recognition in electronic medical records.
\newblock \emph{New York Scientific Data Summit (NYSDS), 2016: IEEE}, 2016.

\end{thebibliography}
\bibliographystyle{icml2020}

\begin{table*}
\centering
\begin{tabular}{lllr}  
\toprule
Flair & Spacy & Stanford & BERT+BiGRU+CRF \\
\midrule
 & \$1–1.5 trillion & daniel & around two percent \\
 & annual & international monetary fund & bank \\
 & around two percent & new york. & capita income \\
 & daniel kaufmann & nrgi & daniel kaufmann \\
 & every year & xinhua & every year \\
 & kaufmann &  & infant mortality \\
 & new york &  & international monetary fund annual meeting \\
 & nrgi &  & natural resource governance \\
 & sunday &  & natural resource governance institute \\
 & the natural resource governance institute nrgi &  & new york \\
 & the world bank group &  & public governance \\
 & xinhua news agency &  & rule \\
 &  &  & the natural resource governance institute nrgi \\
 &  &  & the world bank group \\
 &  &  & xinhua news agency \\
\bottomrule
\end{tabular}
\caption{Recognised Named Entities Per Model - Example 1}
\label{tab:booktabs2}
\end{table*}

\begin{table*}
\centering
\begin{tabular}{lllr}  
\toprule
Flair & Spacy & Stanford & BERT+BiGRU+CRF \\
\midrule
indian & 2013 & aam aadmi & aam aadmi party \\
sheila dikshit & aap & arvind kejriwal. & aap bandwagon \\
 & arvind kejriwal & congress & arvind kejriwal \\
 & congress & ease. & delhi \\
 & delhi & indian & delhi assembly constituency \\
 & first & new delhi & giant killer \\
 & five-year & sheila dikshit & indian revenue service \\
 & indian & sunita kejriwal & political novice \\
 & irs &  & sheila dikshit \\
 & sunita kejriwal &  & sunita kejriwal \\
 & the aam aadmi party's &  & the aam aadmi party \\
 & winter afternoon &  & winter afternoon \\
\bottomrule
\end{tabular}
\caption{Recognised Named Entities Per Model - Example 2}
\label{tab:booktabs3}
\end{table*}

\end{document}